\documentclass{article}

\usepackage{microtype}
\usepackage{graphicx}
\usepackage{subfigure}
\usepackage{booktabs} 
\usepackage{amsfonts} 
\usepackage{amsthm}
\usepackage{amsmath}
\usepackage{multirow,enumitem}
\usepackage{makecell}
\usepackage{hyperref}

\usepackage[accepted]{icml2021}

\icmltitlerunning{GraphDF: A Discrete Flow Model for Molecular Graph Generation}

\begin{document}

\twocolumn[
\icmltitle{GraphDF: A Discrete Flow Model for Molecular Graph Generation}




\begin{icmlauthorlist}
\icmlauthor{Youzhi Luo}{tamu}
\icmlauthor{Keqiang Yan}{tamu}
\icmlauthor{Shuiwang Ji}{tamu}
\end{icmlauthorlist}

\icmlaffiliation{tamu}{Department of Computer Science \& Engineering, Texas A\&M University, TX, USA}

\icmlcorrespondingauthor{Shuiwang Ji}{sji@tamu.edu}

\icmlkeywords{Machine Learning, ICML}

\vskip 0.3in
]



\printAffiliationsAndNotice{}  

\begin{abstract}
We consider the problem of molecular graph generation using deep models. While graphs are discrete, most existing methods use continuous latent variables, resulting in inaccurate modeling of discrete graph structures. In this work, we propose GraphDF, a novel discrete latent variable model for molecular graph generation based on normalizing flow methods. GraphDF uses invertible modulo shift transforms to map discrete latent variables to graph nodes and edges. We show that the use of discrete latent variables reduces computational costs and eliminates the negative effect of dequantization. Comprehensive experimental results show that GraphDF outperforms prior methods on random generation, property optimization, and constrained optimization tasks.
\end{abstract}

\section{Introduction}
\label{sec:intro}

A fundamental problem in drug discovery \cite{stokes2020deep,wang2020moleculekit} and chemistry is to design novel molecules with specific properties. This task remains to be challenging because the space of molecules is naturally discrete and very large, with estimated size on the order of $10^{33}$ \cite{polishchuk2013estimation}. Recent advances in deep generative models have lead to significant progress in this field. Notably, many recent studies represent molecular structures as graphs and propose to generate novel molecular graphs with advanced deep generative models. From a conceptual and computational perspective, these methods first map molecular graphs to vectors in the continuous latent space. When generating, the generative models project a randomly picked continuous vector in the latent space back to the molecular graph space.

To make use of generative models with continuous latent variables, existing methods \cite{madhawa2019graphnvp,shi2020graphaf,zang2020moflow} convert discrete graph data to continuous data by adding real-valued noise. However, such dequantization processing prevents models from capturing the original discrete distribution of discrete graph structures, thus increasing the difficulty of model training. This becomes a key limitation that makes it hard to capture the true distribution of graph structures and generate diverse molecules.

In this work, we propose GraphDF, a generative model using discrete latent variables for molecular graph generation. GraphDF generates molecular graphs by sequentially sampling discrete latent variables and mapping them to new nodes and edges via invertible modulo shift transforms. Compared with previous methods, our proposed discrete transform eliminates the cost of computing Jacobian matrix and overcomes the limitations resulted from dequantization. Experimental results show that GraphDF can outperform previous state-of-the-art methods over various molecule generation tasks. The implementation of GraphDF has been integrated into the DIG \cite{liu2021dig} framework.

\section{Background and Related Work}
\label{sec:related}

\subsection{Molecule Generation}
\label{sec:mol_gen}
Let $\mathcal{M}=\{M_i\}_{i=1}^m$ be a set of molecules, $S(M)\in\mathbb{R}$ be a function computing a specific property score of the molecule $M$, and $\mbox{SIM}(M,M')\in[0,1]$ be a function measuring the similarity between two molecules $M$ and $M'$. The three molecule generation tasks can be defined as below.
\begin{itemize}[leftmargin=*]
    \item Learning a generation model $p_\theta(\cdot)$ from $\mathcal{M}$, where $p_\theta(M)$ is the probability of sampling the molecule $M$ from the model.
    \item Learning a property optimization model $p_\theta(\cdot)$ with respect to $S$ by maximizing $\mathbb{E}_{M \sim p_\theta}[S(M)]$.
    \item Learning a constrained optimization model $p_\theta(\cdot|M)$, so as to maximize $\mathbb{E}_{M'|M \sim p_\theta}[S(M')]$ while satisfying $\mbox{SIM}(M,M')>\delta$, where $M$ is a given molecule, $M'$ is the optimized molecule, and $\delta$ is a similarity threshold.
\end{itemize}

Early attempts for molecule generation~\cite{gomez2018automatic,kusner2017grammar,dai2018syntaxdirected} represented molecules as SMILES strings \cite{weininger1988smiles} and developed sequence generation models. These models may not easily learn the complicated grammatical rules of SMILES and thus could not generate syntactically valid sequences well. More recent studies represent molecules as graphs, and significant progress has been made in molecular graph generation with deep generative models. A variety of molecular graph generation methods have been developed based on variational auto-encoders (VAE) \cite{kingma2013auto}. Some of these methods generate molecular graphs from a tree structure \cite{jin2018junction,kajino2019mhg}; others add a node or an edge sequentially one at a time \cite{liu2018constrained}, or generate node and adjacency matrices directly in one-shot sampling \cite{simonovsky2018graphvae,ma2018constrained}. For VAE-based models, Bayesian optimization can be applied in the latent space to search molecules with optimized property scores \cite{jin2018junction,kajino2019mhg}. Generative adversarial networks (GAN) \cite{goodfellow2014gen} have also been used in some molecular graph generators \cite{de2018molgan,you2018goal}. Similar to SeqGAN \cite{yu2017seqgan}, these models are trained with reinforcement learning.

In addition to methods based on VAE and GAN, normalizing flow models have been used in various generation tasks \cite{dinh2014nice, dinh2016density, danilo2015variational}. These models define invertible transformations between latent variables and data samples, thus allowing for exact likelihood computation. It has been demonstrated that these models are able to capture the density of molecular graphs more accurately \cite{shi2020graphaf,zang2020moflow}. Our proposed discrete flow model is based on the normalizing flow method.

\subsection{Flow Models}

A flow model is a parametric invertible function $f_\theta: \mathbb{R}^d\to\mathbb{R}^d$, defining an invertible mapping from the latent variable $z\in\mathbb{R}^d$ to the data point $x\in\mathbb{R}^d$, where $x\sim p_X(x)$ and $z\sim p_Z(z)$ are random variables. The latent distribution $p_Z$ is a pre-defined probability distribution, \emph{e.g.}, a Gaussian distribution. The data distribution $p_X$ is unknown. But given an arbitrary data point $x\in\mathbb{R}^d$, we can use the change-of-variable theorem to compute its log-likelihood as
\begin{equation}
  \label{eqn:con_flow}
    \log p_X(x)=\log p_Z\left(f_\theta^{-1}(x)\right)+\log\left|\mbox{det}\frac{\partial f_\theta^{-1}(x)}{\partial x}\right|,
\end{equation}
where $\frac{\partial f_\theta^{-1}(x)}{\partial x}\in\mathbb{R}^{d\times d}$ is the Jacobian matrix. To train a flow model $f_\theta$ on a dataset $\mathcal{X}=\{x_i\}_{i=1}^m$, we use Eqn.~(\ref{eqn:con_flow}) to compute the exact log-likelihoods of all data points and update the parameter $\theta$ by maximizing the log-likelihoods via gradient descent. To randomly sample a data point from $f_\theta$, a latent variable $z$ is first sampled from the pre-defined latent distribution $p_Z$. Then we can obtain a data point $x$ by performing the feedforward transformation $x=f_\theta(z)$. Therefore, the mapping $f_\theta$ should be invertible, and the computation of the Jacobian matrix $\frac{\partial f_\theta^{-1}(x)}{\partial x}$ needs to be tractable in order to enabling efficient training and sampling. A common operation satisfying these two requirements is the affine coupling layer \cite{dinh2016density}, which has been used frequently in deep flow models.

Recently, flow models have been used in molecule or graph generation tasks. \citet{liu2019gnf} proposes an invertible message passing neural network, known as GNF, for simple community graph generation. 
Following GraphVAE \cite{simonovsky2018graphvae}, several recent studies generate molecular graphs by generating node and adjacency matrices in a one-shot fashion via flow models, including GraphNVP \cite{madhawa2019graphnvp}, GRF \cite{honda2019grf}, and MoFlow \cite{zang2020moflow}. 
GraphAF \cite{shi2020graphaf} generates molecular graphs by adding nodes and edges sequentially via an autoregressive flow model \cite{papamakarios2017af}. GraphCNF \cite{lippe2021graphcnf} jointly optimizes a variational inference model and a flow model for graph generation.

\section{A Discrete Flow Model for Graphs}
\label{sec:method}

While all current molecular graph generation methods use continuous latent variables, we argue that such continuous prior distribution is not suitable to model molecular graph data, which are naturally discrete. On the other hand, some recent attempt has been made to develop generative models with discrete latent variables \cite{tran2019discrete}. It is, however, only able to generate 1D sequences, such as texts, and cannot be used to generate richly structured objects such as graphs.

In this section, we introduce GraphDF, a novel molecular graph generative model with discrete latent variables. To our best knowledge, GraphDF is the first work which demonstrates that discrete latent variable models have strong capacity to model the density of complicated molecular graph data.

\subsection{An Overview of Sequential Generation}
\label{sec:seq_gen}

Let $k$ be the total number of node types and $c$ be the total number of edge types. We use $G=(X,A)$ to denote a molecular graph with $n$ nodes, where $X\in\{0,1\}^{n\times k}$ is the node matrix and $A\in\{0,1\}^{n\times n\times c}$ is the adjacency tensor. $X[i,u]=1$ represents that the $i$-th node has type $u$, and $A[i,j,v]=1$ represents that there is an edge with type $v$ between the $i$-th node and $j$-th node. If there is no edge between the $i$-th node and $j$-th node, then $A[i,j,v]=0$ for $v=0,\dots,c-1$.

\begin{figure}[!t]
    \vskip 0.2in
    \begin{center}
    \centerline{\includegraphics[width=0.9\columnwidth]{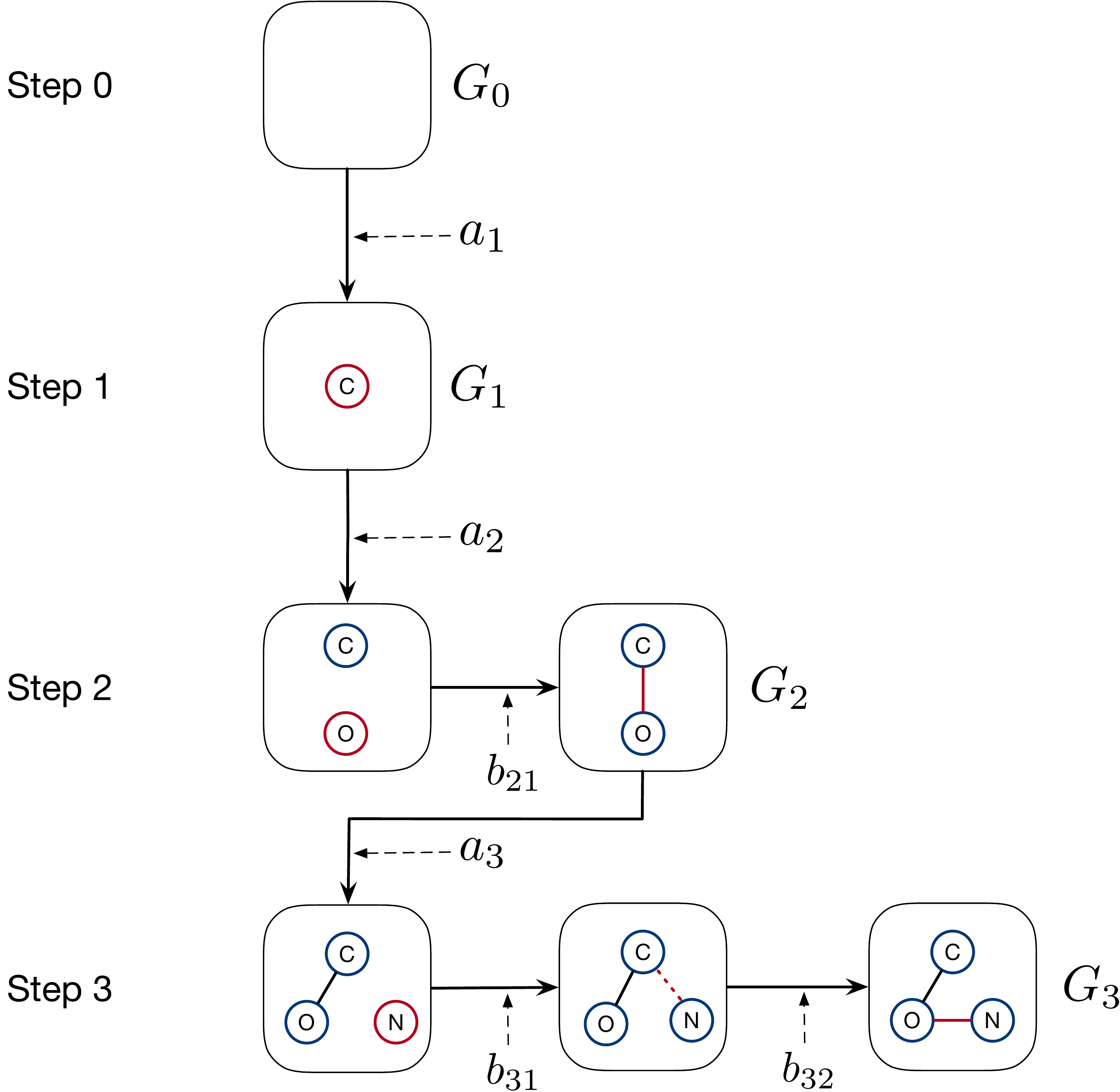}}
    \caption{An illustration of sequential generation procedure.}
    \label{fig:seq_gen}
    \end{center}
    \vskip -0.2in
\end{figure}

Similar to MolecularRNN \cite{popova2019molecularrnn}, we generate molecular graphs by a sequential decision process. We start from an empty molecular graph $G_0$. At step $i$, let the current sub-graph be $G_{i-1}$ which contains $i-1$ nodes. The generative model $f_\theta$ first generates a new node with type $a_i$ based on a sampled latent variable $z_i$, where $a_i\in\{0,\dots,k-1\}$. Afterwards, the edges between the new node and all nodes in $G_{i-1}$ are generated sequentially by $f_\theta$. Specifically, the edge incident to the $j$-th node of $G_{i-1}$ is generated based on the latent variable $z_{ij}$, whose edge type $b_{ij}\in\{0,\dots,c-1,c\}$, where the extra edge type $c$ means no edge. If $b_{ij}$ violates the chemical bond valency rule, it is re-generated by re-sampling a new $z_{ij}$. The above process is an autoregressive function of previously generated elements and repeats until no new edge is added at step $i$, i.e., $b_{ij}=c$ for $j=1,\dots,i-1$. An example of this process is given in Fig.~\ref{fig:seq_gen}. In summary, generating a molecular graph $G$ is equivalent to generating the sequence representation $S_G=(a_1,a_2,b_{21},a_3,b_{31},b_{32},a_4,\dots)$ by
\begin{equation}
    \label{eqn:seq_gen}
    \begin{split}
        a_i&=f_\theta(a_1,\dots,b_{i-1,i-2};z_i),\quad i>0,\\
        b_{ij}&=f_\theta(a_1,\dots,b_{i,j-1};z_{ij}),\quad j=1,\dots,i-1.
    \end{split}
\end{equation}

\subsection{Generation with Discrete Latent Variables}

\begin{figure}[!t]
    \vskip 0.2in
    \begin{center}
    \centerline{\includegraphics[width=0.9\columnwidth]{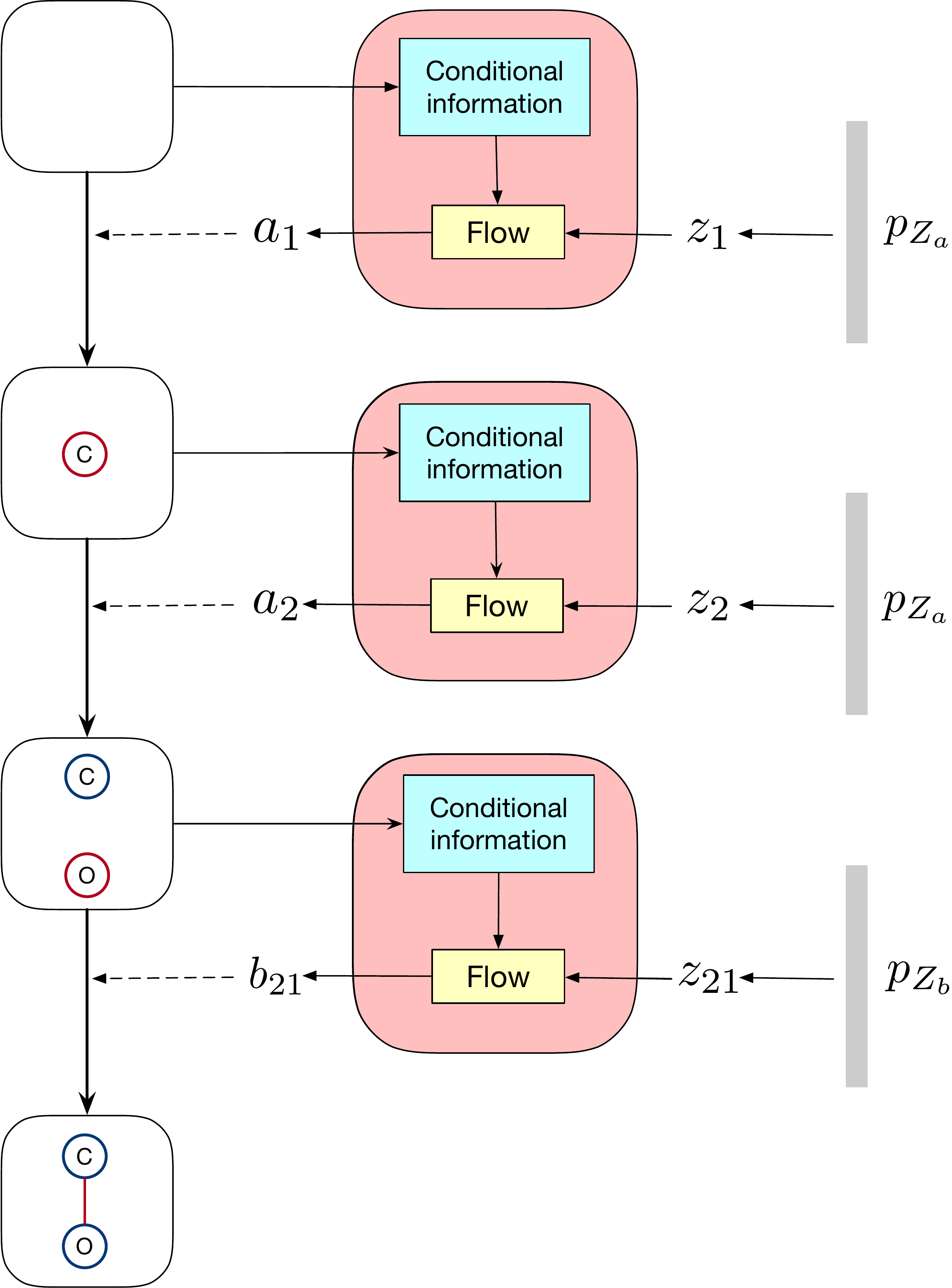}}
    \caption{An illustration of generation with GraphDF framework.}
    \label{fig:graphdf_gen}
    \end{center}
    \vskip -0.2in
\end{figure}

In our method, all latent variables are discrete and sampled from multinomial distributions. Specifically, $z_i$ is sampled from a multinomial distribution $p_{Z_a}$ with trainable parameters $(\alpha_0,\dots,\alpha_{k-1})$ as 
\begin{equation}
    p_{Z_a}(z_i=s)=\frac{\exp{(\alpha_s)}}{\sum_{t=0}^{k-1}\exp{(\alpha_t)}}.
\end{equation}
Similarly, $z_{ij}$ is sampled from the multinomial distribution $p_{Z_b}$ with trainable parameters $(\beta_0,\dots,\beta_{c})$. We use a discrete flow model to reversibly map discrete latent variables to new nodes and edges. The discrete transform used in the discrete flow is a modulo shift transform in the form of
\begin{equation}
    q(z)=(z+\mu)\bmod t,
\end{equation}
where $t$ is the number of categories, and $z,\mu\in\{0,\dots,t-1\}$. Formally, the discrete transform for generating new nodes and edges in Eqn.~(\ref{eqn:seq_gen}) is the composition of $D$ modulo shift modules as
\begin{equation}
\label{eqn:graphdf_gen}
    \begin{split}
        a_i&=q_i^D\circ\dots \circ q_i^1(z_i),\quad i>0,\\
        b_{ij}&=q_{ij}^D\circ\dots \circ q_{ij}^1(z_{ij}),\quad j=1,\dots,i-1,
    \end{split}
\end{equation}
where $f\circ g(\cdot)$ is defined as $f\left(g(\cdot)\right)$, and
\begin{equation}
    \begin{split}
        q_i^d(z)&=(z+\mu_i^d)\bmod k,\quad d=1,\dots,D\\
        q_{ij}^d(z)&=(z+\mu^d_{ij})\bmod (c+1),\quad d=1,\dots,D.
    \end{split}
\end{equation}
Here the shift factors $\mu_i^d$ and $\mu_{ij}^d$ are functions of $(a_1,a_2,b_{21},\dots,b_{i-1,i-2})$ and $(a_1,a_2,b_{21},\dots,b_{i,j-1})$, respectively. They both play the role of capturing conditional information from previously generated elements. The calculation of $\mu_i^d$ and $\mu_{ij}^d$ will be described in Sec.~\ref{sec:con_info}. An illustrative example of generating a simple molecular graph with this framework is given in Fig.~\ref{fig:graphdf_gen}.

The use of discrete latent variables is the key factor distinguishing our method from other generation methods based on flow models. Our model has several advantages. The change-of-variable formula of discrete invertible mapping does not involve the Jacobian matrix term in Eqn.~(\ref{eqn:con_flow}). Hence the probabilities of sampled nodes and edges can be directly obtained as 
\begin{equation}
\label{eqn:action_prob}
    \begin{split}
        p(a_i|a_1,a_2,b_{21},\dots,b_{i-1,i-2})&=p_{Z_a}(z_i),\\
        p(b_{ij}|a_1,a_2,b_{21},\dots,b_{i,j-1})&=p_{Z_b}(z_{ij}),
    \end{split}
\end{equation}
leading to dramatic reduction in computational cost. 

More importantly, discrete latent variables free the generative model from the defects of dequantization. Previous methods \cite{madhawa2019graphnvp, shi2020graphaf, zang2020moflow} convert categorical representation of graph nodes and edges to continuous data by adding real-valued noise to their one-hot encodings. Then continuous flow models map such pseudo continuous data to latent variables subject to Gaussian distribution. We argue that such a dequantization operation can be problematic. First, the generative model does not capture the true distribution of original discrete data, but the distorted distribution after dequantization. Second, due to the randomness of noise, the resulting continuous data points can be very different even if they are obtained by performing dequantization on the same graph. Hence during training, a single graph may be mapped to two distant points in the latent space, leading to model oscillation and failure to converge. However, our GraphDF maps the discrete graph data to discrete latent space by discrete invertible flow without introducing any distribution distortion. Hence it can model the underlying distribution of graph structures more accurately, thereby eliminating the impact of two issues caused by dequantization.

Note that our method is fundamentally different from GraphCNF \cite{lippe2021graphcnf}, though dequantization is avoided in both methods. GraphCNF maps discrete values to continuous latent space where different categories occupy non-overlapping regions. Such a discrete-to-continuous mapping leads to distribution distortions. In the generation process, discrete values are sampled by applying Bayes, and all graph nodes and edges are generated at once. On the other hand, GraphDF uses a fundamentally different approach to model the discrete distribution. GraphDF completely discards continuous latent space and maps discrete values to discrete latent space by the modulo shift transform. No distribution distortion is introduced here. In addition, compared with GraphCNF, the sequential fashion of GraphDF can capture the graph structure information efficiently and guarantee the chemical validity because of using bond valency check in the generation.

\subsection{Inference and Training}

The inverse process of generation is inference, which involves mapping from graphs to discrete latent variables. Given a graph $G=(X,A)$, we compute its sequence representation $S_G$ by re-ordering its nodes and edges according to the breadth-first search (BFS) order. Then we can obtain the discrete latent variables corresponding to each element of this sequence by inverting Eqn.~(\ref{eqn:graphdf_gen}) as
\begin{equation}
    \label{eqn:flow_infer}
    \begin{split}
        z_i&=o_i^1\circ\dots \circ o_i^D(a_i),\quad i>0,\\
        z_{ij}&=o_{ij}^1\circ\dots \circ o_{ij}^D(b_{ij}),\quad j=1,\dots,i-1,
    \end{split}
\end{equation}
where
\begin{equation}
    \begin{split}
        o_i^d(z)&=(z-\mu_i^d)\bmod k,\quad d=1,\dots,D\\
        o_{ij}^d(z)&=(z-\mu^d_{ij})\bmod (c+1),\quad d=1,\dots,D.
    \end{split}
\end{equation}
Similar to the generation process, the shift factors $\mu_i^d$ and $\mu_{ij}^d$ capture autoregressive information from elements in $S_G$ before $a_i$ and $b_{ij}$, respectively.

We train a GraphDF model on a molecular graph dataset by maximizing the log-likelihoods of all data samples. Given a molecular graph $G$, we first perform inference on $G$ to compute the latent variables corresponding to its nodes and edges. Then the log-likelihood of $G$ can be computed as
\begin{equation}
\hspace{-0.3cm}\log p(G)=\sum_{i=1}^n\left(\log p_{Z_a}(z_i)+\sum_{j=1}^{i-1}\log p_{Z_b}(z_{ij})\right).
\end{equation}
The parameters of GraphDF models and discrete prior distributions $p_{Z_a},p_{Z_b}$ are updated by gradient descent to maximize the log-likelihoods of all data samples.

\subsection{Invariant and Discrete Conditional Generation}
\label{sec:con_info}

The transformation of discrete flow in Eqn.~(\ref{eqn:graphdf_gen}) is autoregressive and requires capturing conditional information into $\mu_i^d$ and $\mu_{ij}^d$. \citet{tran2019discrete} uses language models, like LSTM \cite{hochreiter1997long} and Transformer \cite{vaswani2017attention}, in discrete flow models to develop sequence generators. However, the same method cannot be used directly to generate $S_G$ for two reasons as below.
\begin{itemize}[leftmargin=*]
    \item Language models are only aware of the node and edge type information encoded in $S_G$, while rich structure information of graphs is not captured.
    \item There is no intrinsic order in an intermediately generated sub-graph, but language models assume an order when dealing with $S_G$. For instance, for $(a_1,a_2)=(1,0)$ and $(a_1,a_2)=(0,1)$, the generative model should give the same estimate $p(b_{21}|a_1,a_2)$, but language models cannot provide such a guarantee.
\end{itemize}

To tackle the two issues, we propose to capture conditional information from intermediate sub-graphs by using graph convolutional networks (GCN) \cite{kipf2017semi,gao18large,gao19graph}. Specifically, let the sub-graph generated immediately before adding the node $a_i$ be $G_{i-1}=(X,A)$, we use an $L$-layer relational GCN (R-GCN) \cite{Schlichtkrull2018rgcn} to obtain node embeddings $H^L\in\mathbb{R}^{n\times r}$ from $G_{i-1}$ by performing $L$ message passing operations. The $\ell$-th message passing is performed by:
\begin{equation}
    H^{\ell} = \sum_{v=1}^c\mbox{ReLU}\left( \tilde{D}_{v}^{-\frac{1}{2}}\tilde{A}_{v}\tilde{D}_{v}^{-\frac{1}{2}}H^{\ell-1}W_v^{\ell} \right),
\end{equation}
where $H^0=X$, $\tilde{A}_{v}=A[:,:,v]+I$, $\tilde{D}_{v}$ is a diagonal matrix and  $\tilde{D}_{v}[j,j]=\sum\limits_{u=1}^n\tilde{A}_{v}[j,u]$, $\{W_v^{\ell}\}_{v=1}^c$ are trainable weight matrices of the $\ell$-th layer. Note that the same R-GCN is shared among all modulo shift modules in the discrete flow. Afterwards, $\mu_i^d$ is calculated by a classification network based on multi-layer perceptrons (MLP) as
\begin{equation}
\label{eqn:node_shift}
    \begin{split}
        h&=\mbox{sum}(H^L),\\
        \mu_i^d&=\arg\max \mbox{MLP}^d_a\left(h\right),\quad d=1,\dots,D,
    \end{split}
\end{equation}
where $h\in\mathbb{R}^r$ is the graph embedding obtained by performing sum-pooling on node embeddings $H^L$, $\mbox{MLP}^d_a$ projects the input to a $k$-dimensional logits vector, and $\arg\max$ outputs the index position of the maximum logits value. Since $\arg\max$ is not differentiable, we use the softmax-temperature function to approximate the gradient \cite{jang2016categorical} during training as 
\begin{equation}
    \begin{split}
        \gamma^d&=\mbox{MLP}_a^d(h),\quad d=1,\dots,D,\\
        \frac{\mathrm{d} \mu_d}{\mathrm{d} \gamma^d}&\approx \frac{\mathrm{d}}{\mathrm{d} \gamma^d}\mbox{softmax}(\frac{\gamma^d}{\tau}),\quad d=1,\dots,D,
    \end{split}
\end{equation}
where we fix the temperature $\tau=0.1$ in experiments.

Similarly, to obtain $\mu_{ij}^d$, we use the R-GCN to extract node embeddings $H^L$ and graph embedding $h$ from $G_{i-1,j}$, which is the sub-graph generated immediately before adding the edge $b_{ij}$. Then we compute $\mu_{ij}^d$ by MLP as:
\begin{equation*}
\label{eqn:edge_shift}
    \mu_{ij}^d=\arg\max \mbox{MLP}_b^d\left(\mbox{Con}(h,H_i^L,H_j^L)\right),d=1,\dots,D,
\end{equation*}
where $H_i^L$ and $H_j^L$ denote the node embeddings of the $i$-th and $j$-th node of $G_{i-1,j}$, respectively, and $\mbox{Con}(\cdot)$ is the concatenation operation. The softmax approximation is also used here to perform back propagation through $\arg\max$.

By using our proposed method, the discrete flow model can elegantly extract conditional information from graph structures, which is then incorporated into the mapping from discrete latent space to newly generated graph nodes and edges. The conditional information is guaranteed to be invariant to the order of previously generated elements. We summarize the detailed generation and training algorithms of GraphDF in Appendix~\ref{app:gen_alg} and~\ref{app:train_alg}, respectively.

\subsection{Property Optimization with Reinforcement Learning}
\label{sec:rl}

The above discussions involve molecular graph generation independent of property scores.
However, both property optimization and constrained optimization require the generated molecules to have optimal property scores. To this end, we use reinforcement learning to fine-tune a trained GraphDF model to optimize the specified property score.

We can formulate the sequential generation process with GraphDF as a Markov Decision Process (MDP). Formally, at each step, we treat the currently generated sub-graph as the state, and adding a new node or edge as the action. The action probabilities $p(a_i|G_{i-1})$ and $p(b_{ij}|G_{i-1,j})$ are thus determined by the GraphDF model in Eqn.~(\ref{eqn:action_prob}). The rewards consist of intermediate reward and final reward. The intermediate reward is the penalty for violating chemical bond valency rule. The final reward incorporates the property score and penalty for containing excessive steric strain or violating functional group filters of ZINC \cite{irwin2012zinc} in the finally generated molecule.

Following GCPN \cite{you2018goal}, we use Proximal Policy Optimization (PPO) \cite{schulman2017proximal}, a powerful policy optimization algorithm, to fine-tune GraphDF models in the above reinforcement learning environment. The loss function for a generated graph $G$ is
\begin{equation}
\begin{split}
    L(G,\theta)&=\sum_{i=1}^n \left(L^{\mbox{c}}(r_i,A_i)+\sum_{j=1}^n L^{\mbox{c}}(r_{ij},A_{ij})\right),\\
    L^{\mbox{c}}(r,A)&=\min\{rA, \mbox{clip}(r,1-\epsilon,1+\epsilon)A\},
\end{split}
\end{equation}
where $r_i=\frac{p(a_i|G_{i-1};\theta)}{p(a_i|G_{i-1};\theta_{\text{old}})}$ and $r_{ij}=\frac{p(b_{ij}|G_{i-1,j};\theta)}{p(b_{ij}|G_{i-1,j};\theta_{\text{old}})}$ are ratios of action probabilities by new policy and old policy. $A_i$ and $A_{ij}$ are estimated advantage functions, for which we use the accumulated rewards of future steps in experiments.

\section{Experiments}
\label{exp}

In this section, we evaluate our GraphDF on three tasks of molecule generation, as mentioned in Section~\ref{sec:mol_gen}. We show that over most tasks, GraphDF can outperform three strong baselines, JT-VAE \cite{jin2018junction}, GCPN \cite{you2018goal} and MRNN \cite{popova2019molecularrnn}, and other molecular graph generation methods based on flow models including GraphNVP \cite{madhawa2019graphnvp}, GRF \cite{honda2019grf}, GraphAF \cite{shi2020graphaf}, MoFlow \cite{zang2020moflow} and GraphCNF \cite{lippe2021graphcnf}.

\subsection{Random Generation}

\begin{table*}[!t]
    \caption{Random generation performance on ZINC250K dataset. Results with $\ast$ are taken from \citet{shi2020graphaf}. Results of MoFlow (with $\diamond$) are obtained by running its official source code.}
    \label{tab:zinc250k}
    \vskip 0.15in
    \begin{center}
    \begin{small}
    \begin{tabular}{lccccc}
        \toprule
        Method & Validity & Validity w/o check & Uniqueness & Novelty & Reconstruction \\
        \midrule
        JT-VAE & \textbf{100\%} & n/a & \textbf{100\%}$^\ast$ & \textbf{100\%}$^\ast$ & 76.7\% \\
        GCPN & \textbf{100\%} & 20\%$^\ast$ & 99.97\%$^\ast$ & \textbf{100\%}$^\ast$ & n/a \\
        MRNN & \textbf{100\%} & 65\% & 99.89\% & \textbf{100\%} & n/a \\
        GraphNVP & 42.6\% & n/a & 94.8\% & \textbf{100\%} & \textbf{100\%} \\
        GRF & 73.4\% & n/a & 53.7\% & \textbf{100\%} & \textbf{100\%} \\
        GraphAF & \textbf{100\%} & 68\% & 99.1\% & \textbf{100\%} & \textbf{100\%} \\
        MoFlow & \textbf{100\%}$^\diamond$ & 50.3\%$^\diamond$ & 99.99\%$^\diamond$ & \textbf{100\%}$^\diamond$ & \textbf{100\%}$^\diamond$ \\
        GraphCNF & 96.35\% & n/a & 99.98\% & 99.98\% & \textbf{100\%} \\
        \midrule
        GraphDF & \textbf{100\%} & \textbf{89.03\%} & 99.16\% & \textbf{100\%} & \textbf{100\%} \\
        \bottomrule
    \end{tabular}
    \end{small}
    \end{center}
    \vskip -0.1in
\end{table*}

\begin{table*}[!t]
    \caption{Random generation performance on QM9 dataset. Results of MoFlow (with $\diamond$) are obtained by running its official source code.}
    \label{tab:qm9}
    \vskip 0.15in
    \begin{center}
    \begin{small}
    \begin{tabular}{lccccc}
        \toprule
        Method & Validity & Validity w/o check & Uniqueness & Novelty & Reconstruction \\
        \midrule
        GraphNVP & 83.1\% & n/a & \textbf{99.2\%} & 58.2\% & \textbf{100\%} \\
        GRF & 84.5\% & n/a & 66\% & 58.6\% & \textbf{100\%} \\
        GraphAF & \textbf{100\%} & 67\% & 94.15\% & 88.83\% & \textbf{100\%} \\
        MoFlow & \textbf{100\%}$^\diamond$ & \textbf{88.96\%}$^\diamond$ & 98.53\%$^\diamond$ & 96.04\%$^\diamond$ & \textbf{100\%}$^\diamond$ \\
        \midrule
        GraphDF & \textbf{100\%} & 82.67\% & 97.62\% & \textbf{98.1\%} & \textbf{100\%} \\
        \bottomrule
    \end{tabular}
    \end{small}
    \end{center}
    \vskip -0.1in
\end{table*}

\begin{table*}[!t]
    \caption{Random generation performance on MOSES dataset. Results with $\ast$ are taken from \citet{polykovskiy2020moses}.}
    \label{tab:moses}
    \vskip 0.15in
    \begin{center}
    \begin{small}
    \begin{tabular}{lccccc}
        \toprule
        Method & Validity & Validity w/o check & Uniqueness & Novelty & Reconstruction \\
        \midrule
        JT-VAE & \textbf{100\%}$^\ast$ & n/a & 99.96\%$^\ast$ & 91.43\%$^\ast$ & n/a \\
        GraphCNF & 95.66\% & n/a & 99.98\% & \textbf{100\%} & n/a \\
        GraphAF & \textbf{100\%} & 71\% & \textbf{99.99\%} & \textbf{100\%} & \textbf{100\%} \\
        \midrule
        GraphDF & \textbf{100\%} & \textbf{87.58\%} & 99.55\% & \textbf{100\%} & \textbf{100\%} \\
        \bottomrule
    \end{tabular}
    \end{small}
    \end{center}
    \vskip -0.1in
\end{table*}

\begin{table*}[!t]
    \caption{Random generation performance on COMMUNITY-SMALL and EGO-SMALL datasets. Following GNF \cite{liu2019gnf}, we evaluate with MMD metric, the first set of results are obtained using node distribution matching, and the second set of results are obtained from evaluating 1024 generated graphs without node distribution matching.}
    \label{tab:general}
    \vskip 0.15in
    \begin{center}
    \begin{small}
    \begin{tabular}{lcccccc}
       \toprule
       \multirow{2}{*}{Method} & \multicolumn{3}{c}{COMMUNITY-SMALL} & \multicolumn{3}{c}{EGO-SMALL} \\
         & Degree & Cluster & Orbit & Degree & Cluster & Orbit \\
       \midrule
       GNF & 0.20 & 0.20 & 0.11 & 0.03 & \textbf{0.11} & \textbf{0.001} \\
       GraphAF & 0.18 & 0.20 & \textbf{0.02} & \textbf{0.03} & \textbf{0.11} & \textbf{0.001} \\
       GraphDF & \textbf{0.06} & \textbf{0.12} & 0.03 & 0.04 & 0.13 & 0.010 \\
       \midrule
       GNF(1024) & 0.12 & 0.15 & 0.02 & \textbf{0.01} & \textbf{0.03} & \textbf{0.0008} \\
       GraphAF(1024) & 0.06 & \textbf{0.10} & \textbf{0.015} & 0.04 & 0.04 & 0.008 \\
       GraphDF(1024) & \textbf{0.04} & \textbf{0.10} & 0.018 & 0.07 & 0.10 & 0.014 \\
       \bottomrule
    \end{tabular}
    \end{small}
    \end{center}
    \vskip -0.1in
\end{table*}

\textbf{Data.} For random generation of molecular graphs, we train and evaluate GraphDF on three molecule datasets, ZINC250K \cite{irwin2012zinc}, QM9 \cite{ramakrishnan2014quantum}, and MOSES \cite{polykovskiy2020moses}. All molecules are preprocessed to kekulized form by RDKit \cite{rdkit}. In addition, we conduct experiments to show that GraphDF is very general and can be used to graphs other than molecular graphs. Following the experiments of GNF \cite{liu2019gnf}, we test GraphDF on two generic graph datasets, COMMUNITY-SMALL and EGO-SMALL. Detailed information about all datasets used in random generation experiments is provided in Appendix~\ref{app:data}.

\textbf{Setup.} We adopt five widely used metrics to evaluate GraphDF on molecular graph generation. Among all generated molecules, \textbf{Validity} is the percentage of molecules which do not violate chemical valency rule. Since we use re-sampling when valency check fails as mentioned in Sec.~\ref{sec:seq_gen}, we also report \textbf{Validity w/o check}, which is the validity percentage when this valency correction is disabled. Among all generated valid molecules, \textbf{Uniqueness} and \textbf{Novelty} are the percentage of unique molecules and molecules not appearing in the training data, respectively. \textbf{Reconstruction} is the percentage of molecules which can be reconstructed from their latent variables. The five metrics are calculated from 10,000 generated molecular graphs. We find that the reported metrics in MoFlow paper \cite{zang2020moflow} are not calculated from the same 10,000 generated molecules, so we recalculate the metrics of MoFlow by rerunning its official source code for fair comparison. As for COMMUNITY-SMALL and EGO-SMALL datasets, we follow GNF \cite{liu2019gnf} to calculate the Maximum Mean Discrepancy (MMD) \cite{gretton2012kernel} distance of statistics of degrees, clustering coefficients and orbit counts between generated graphs and graphs in the dataset. We calculate the MMD from 1024 generated graphs in the two cases of using and not using node distribution matching.

The flow model in GraphDF consists of 12 modulo shift modules, where the shared R-GCN has 3 message passing layers. The hidden dimension and output dimension of node embeddings are 128. For molecular graphs, the input node embeddings are one-hot embeddings of node types. For graphs in COMMUNITY-SMALL and EGO-SMALL datasets, the input node embeddings are one-hot indicator vectors. After message passing layers, a batch normalization layer and a sum pooling layer are used to get the graph embedding. The two MLPs for calculating $\mu_i^d$ and $\mu_{ij}^d$ both have two fully-connected layers, with a hidden dimension of 128 and a non-linear activation function of tanh. A single R-GCN is used among all 12 modulo shift modules in the discrete flow model, while each modulo shift module has an MLP to calculate $\mu_i^d$ or $\mu_{ij}^d$. We use the above model architecture configuration for all experiments in this paper.

Adam \cite{kingma2015adam} optimizer is used to train GraphDF models. On each molecule dataset, GraphDF is trained for 10 epochs, where the batch size is 32 and the learning rate is 0.001. On COMMUNITY-SMALL and EGO-SMALL datasets, GraphDF is trained for 1000 epochs, where the batch size is 16 and the learning rate is 0.001. See Appendix~\ref{app:exp_detail} for more training and generation details.

\textbf{Results.} Table~\ref{tab:zinc250k}, Table~\ref{tab:qm9} and Table~\ref{tab:moses} show that GraphDF can outperform the state-of-the-art methods over most metrics on all three molecule datasets. Thanks to the invertible property of flow models and valency correction, GraphDF can always achieve 100\% validity and reconstruction rate. In addition, GraphDF achieves reasonably high uniqueness and novelty rate on three datasets, showing that it does not overfit or produce mode collapse. Particularly, GraphDF can outperform GraphAF by a large margin in terms of validity w/o check. Since GraphAF generates molecular graphs in the same sequential fashion but uses continuous flow, we believe that the good performance of GraphDF comes from the usage of discrete flow. It strongly demonstrates that discrete latent variables can learn chemical rules and model the underlying distributions more accurately than continuous latent variables. Besides, results in Table~\ref{tab:general} show that GraphDF can consistently achieve good performance on generic graph data when compared with GNF and GraphAF.

\subsection{Property Optimization}

\begin{table}[!t]
    \caption{Property optimization performance evaluated by top-3 property scores. The scores of MRNN are taken from \citet{shi2020graphaf}.}
    \label{tab:po}
    \vskip 0.15in
    \begin{center}
    \begin{small}
    \begin{tabular}{lcccccc}
       \toprule
       \multirow{2}{*}{Method} & \multicolumn{3}{c}{Penalized logP} & \multicolumn{3}{c}{QED} \\
         & 1st & 2nd & 3rd & 1st & 2nd & 3rd \\
       \midrule
       \makecell[c]{ZINC\\(Dataset)} & 4.52 & 4.3 & 4.23 & 0.948 & 0.948 & 0.948 \\
       \midrule
       JT-VAE & 5.3 & 4.93 & 4.49 & 0.925 & 0.911 & 0.910 \\
       GCPN & 7.98 & 7.85 & 7.80 & \textbf{0.948} & 0.947 & 0.946 \\
       MRNN & 8.63 & 6.08 & 4.73 & 0.844 & 0.796 & 0.736 \\
       GraphAF & 12.23 & 11.29 & 11.05 & \textbf{0.948} & \textbf{0.948} & 0.947 \\
       MoFlow & n/a & n/a & n/a & \textbf{0.948} & \textbf{0.948} & \textbf{0.948} \\
       \midrule
       GraphDF & \textbf{13.7} & \textbf{13.18} & \textbf{13.17} & \textbf{0.948} & \textbf{0.948} & \textbf{0.948} \\
       \bottomrule
    \end{tabular}
    \end{small}
    \end{center}
    \vskip -0.1in
\end{table}

\begin{figure}[!t]
    \vskip 0.2in
    \begin{center}
    \subfigure[Four molecules with top penalized logP scores. Note that they are all long carbon chains, only differing in one atom. These are typical chemical structures with high penalized logP property scores.]{
       \centerline{\includegraphics[width=0.9\columnwidth]{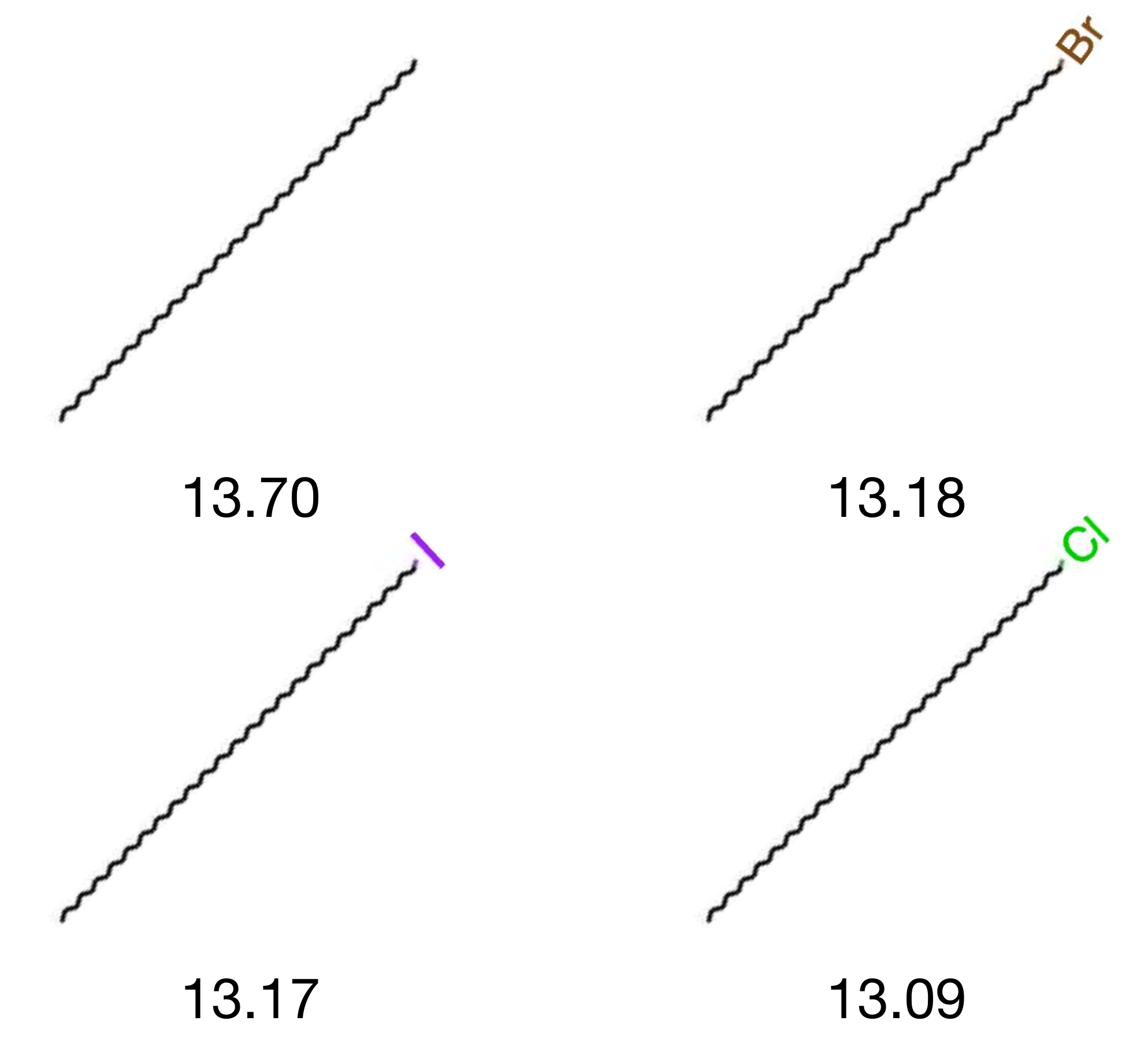}}
    }
    \subfigure[Four molecules with top QED scores.]{
       \centerline{\includegraphics[width=0.9\columnwidth]{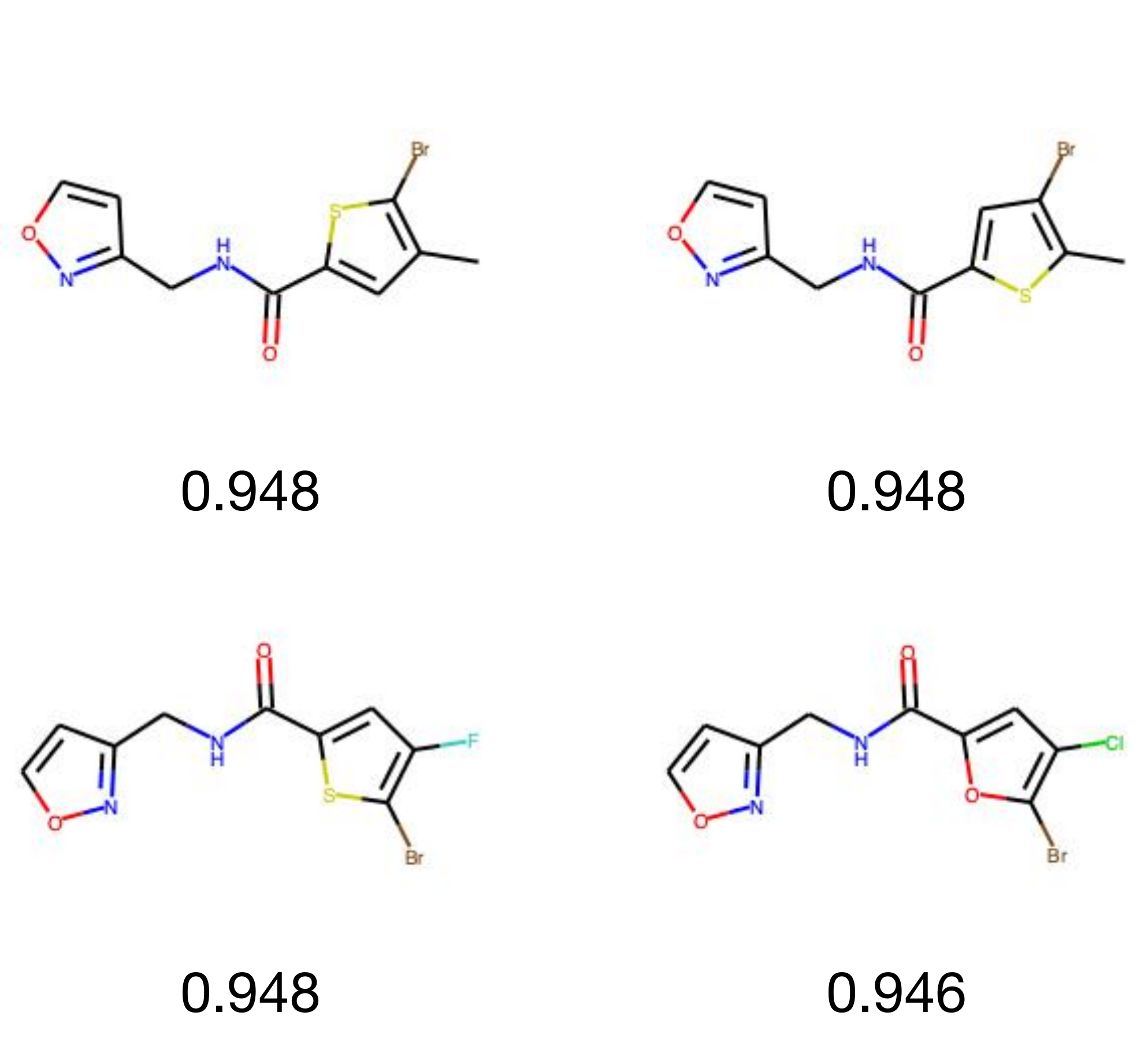}}
    }
    \caption{An illustration of generated molecules with best property scores after property optimization.}
    \label{fig:po}
    \end{center}
    \vskip -0.2in
\end{figure}

\textbf{Setup.} In property optimization task, our objective is to generate novel molecules with high property scores. We use penalized logP and QED \cite{bickerton2012quantifying} as the optimization targets. Penalized logP is the logP score reduced by synthetic accessibility and ring size, and QED is quantitative estimation of drug-likeness. We calculate them using the scripts from the official implementation of GraphAF \cite{shi2020graphaf}.

We use reinforcement learning to optimize these two targets. In this process, both intermediate reward and final reward are considered. The intermediate reward is the penalization for violating chemical bond valency rule. At a generation step, if the action in this step is to add a new edge to the sub-graph and the new edge results in the violation of chemical bond valency rule, then a negative reward of -1 is assigned to this step. Besides, when a complete molecular graph $G$ is generated, the final reward $R(G)$ is calculated as
\begin{equation}
\label{eqn:final_reward}
    R(G)=R_{p}(G)-R_{ss}(G)-R_{f}(G).
\end{equation}
$R_{ss}(G)$ is 1 if $G$ is too sterically strained, i.e., the average angle bend energy of $G$ exceeds 0.82 kcal/mol, otherwise $R_{ss}(G)$ is 0. $R_{f}(G)$ is 1 when functional group filters of ZINC \cite{irwin2012zinc} detect problematic functional groups from $G$, otherwise $R_f(G)$ is 0. $R_{p}(G)$ is related to the property score of $G$. Denote by $\mbox{logP}(\cdot)$ the computation of penalized logP property and $\mbox{QED}(\cdot)$ the computation of QED property, we use $$R_{p}(G)=\exp{\left(\mbox{logP}(G)/3.0-4.0\right)},$$
and $$R_{p}(G)=2.0*\mbox{QED}(G),$$
respectively. The final reward $R(G)$ is distributed to all intermediate steps with a discount factor $\gamma=0.9$. Specifically, if the total number of steps to generate $G$ is $T$, then the assigned reward is $\gamma^{T-t}R(G)$ for the $t$-th step where $t=1,\dots,T$.

We first pretrain GraphDF model on ZINC250K dataset for 1000 epochs and then fine-tune the model with reinforcement learning for 200 iterations. More details about model training are presented in Appendix~\ref{app:exp_detail}. Following previous literatures, we report the top-3 property scores of molecules generated from the model.

\textbf{Results.} We summarize the top-3 property scores of ZINC250K dataset and generation methods in Table~\ref{tab:po}. It shows that GraphDF can outperform all baselines in penalized logP optimization task and achieve comparable performance in QED optimization task. Note that similar reinforcement learning procedure described in Sec.~\ref{sec:rl} is used to optimize the property scores in GCPN \cite{you2018goal} and GraphAF \cite{shi2020graphaf}. Hence, we believe that the good performance of GraphDF demonstrates its strong capacity to explore the complicated chemical structure space and generate diverse molecular graphs. Some generated molecules with high penalized logP scores and QED scores are visualized in Figure~\ref{fig:po}.

\subsection{Constrained Optimization}

\begin{table*}[!t]
    \caption{Constrained optimization performance on 800 molecules used in JT-VAE and GCPN.}
    \label{tab:co_jtvae}
    \vskip 0.15in
    \begin{center}
    \begin{small}
    \begin{tabular}{cccccccccc}
       \toprule
       \multirow{2}{*}{$\delta$} & \multicolumn{3}{c}{JT-VAE} & \multicolumn{3}{c}{GCPN} & \multicolumn{3}{c}{GraphDF}\\
         & Improvement & Similarity & Success & Improvement & Similarity & Success & Improvement & Similarity & Success \\
       \midrule
       0.0 & 1.91$\pm$2.04 & 0.28$\pm$0.15 & 97.5\% & 4.20$\pm$1.28 & 0.32$\pm$0.12 & 100\% & \textbf{5.93$\pm$1.97} & 0.30$\pm$0.12 & 100\% \\
       0.2 & 1.68$\pm$1.85 & 0.33$\pm$0.13 & 97.1\% & 4.12$\pm$1.19 & 0.34$\pm$0.11 & 100\% & \textbf{5.62$\pm$1.65} & 0.32$\pm$0.10 & 100\% \\
       0.4 & 0.84$\pm$1.45 & 0.51$\pm$0.10 & 83.6\% & 2.49$\pm$1.30 & 0.47$\pm$0.08 & 100\% & \textbf{4.13$\pm$1.41} & 0.47$\pm$0.07 & 100\% \\
       0.6 & 0.21$\pm$0.71 & 0.69$\pm$0.06 & 46.4\% & 0.79$\pm$0.63 & 0.68$\pm$0.08 & 100\% & \textbf{1.72$\pm$1.15} & 0.67$\pm$0.06 & 93\% \\
       \bottomrule
    \end{tabular}
    \end{small}
    \end{center}
    \vskip -0.1in
\end{table*}

\begin{table*}[!t]
    \caption{Constrained optimization performance on 800 molecules used in GraphAF and MoFlow.}
    \label{tab:co_graphaf}
    \vskip 0.15in
    \begin{center}
    \begin{small}
    \begin{tabular}{cccccccccc}
       \toprule
       \multirow{2}{*}{$\delta$} & \multicolumn{3}{c}{GraphAF} & \multicolumn{3}{c}{MoFlow} & \multicolumn{3}{c}{GraphDF}\\
         & Improvement & Similarity & Success & Improvement & Similarity & Success & Improvement & Similarity & Success \\
       \midrule
       0.0 & 13.13$\pm$6.89 & 0.29$\pm$0.15 & 100\% & 8.61$\pm$5.44 & 0.30$\pm$0.20 & 98.88\% & \textbf{14.15$\pm$6.86} & 0.29$\pm$0.13 & 100\% \\
       0.2 & 11.90$\pm$6.86 & 0.33$\pm$0.12 & 100\% & 7.06$\pm$5.04 & 0.43$\pm$0.20 & 96.75\% & \textbf{12.77$\pm$6.59} & 0.32$\pm$0.11 & 100\% \\
       0.4 & 8.21$\pm$6.51 & 0.49$\pm$0.09 & 99.88\% & 4.71$\pm$4.55 & 0.61$\pm$0.18 & 85.75\% & \textbf{9.19$\pm$6.43} & 0.48$\pm$0.08 & 99.63\% \\
       0.6 & \textbf{4.98$\pm$6.49} & 0.66$\pm$0.05 & 96.88\% & 2.10$\pm$2.86 & 0.79$\pm$0.14 & 58.25\% & 4.51$\pm$5.80 & 0.65$\pm$0.05 & 92.13\% \\
       \bottomrule
    \end{tabular}
    \end{small}
    \end{center}
    \vskip -0.1in
\end{table*}

\begin{figure}[!t]
    \vskip 0.2in
    \begin{center}
    \centerline{\includegraphics[width=0.9\columnwidth]{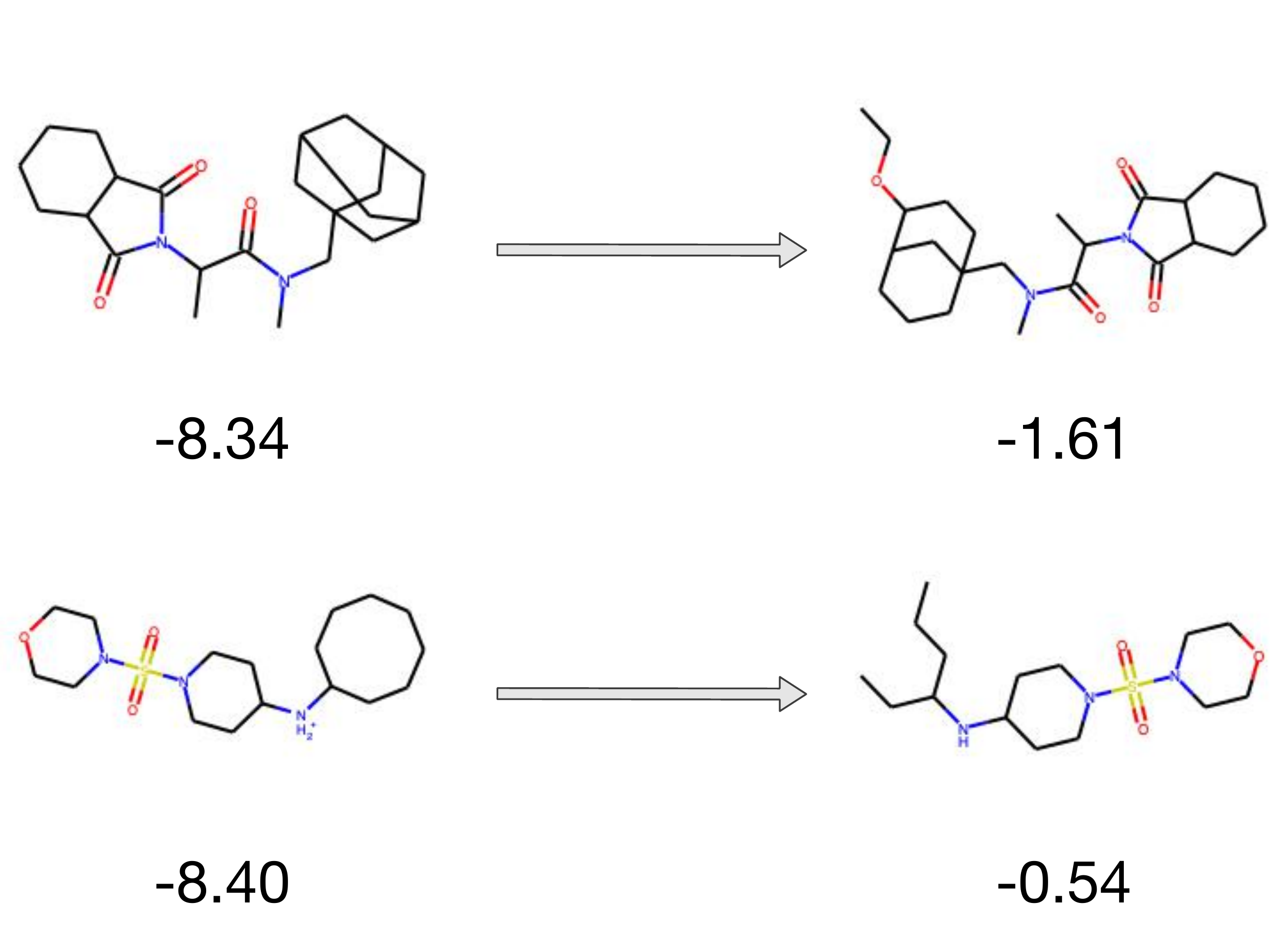}}
    \caption{An illustration of two examples in constrained optimization of penalized logP.}
    \label{fig:co}
    \end{center}
    \vskip -0.2in
\end{figure}

\textbf{Data.} In this task, we aim to modify the input molecular graph so as to improve its penalized logP score while the similarity between the input and modified molecules is higher than a threshold $\delta$. Following JT-VAE \cite{jin2018junction}, GCPN \cite{you2018goal}, GraphAF \cite{shi2020graphaf}, and MoFlow \cite{zang2020moflow}, we use selected 800 molecules from ZINC250K with low penalized logP scores as the input molecules to be optimized. However, we find that the set of 800 molecules used in JT-VAE and GCPN are different from that used in GraphAF and MoFlow, and the official implementation of GCPN does not provide code for re-running constrained optimization on new datasets. Hence, for fair comparison, we train GraphDF and report results on  two datasets, one is used in JT-VAE and GCPN while the other is used in GraphAF and MoFlow.

\textbf{Setup.} Similar to the property optimization task, GraphDF is pretrained on ZINC250K dataset for 1000 epochs and then fine-tuned with reinforcement learning. More details about model training can be found in Appendix~\ref{app:exp_detail}.

We use the reward defined in Eqn.~(\ref{eqn:final_reward}), where $R_p(G)$ is defined as $$R_p(G)=\mbox{logP}(G)-\mbox{logP}(G_{in}),$$ where $G_{in}$ is the input molecular graph. The initial state of a generation process is set to be a sub-graph of the input molecular graph. Specifically, given an input molecular graph $G_{in}$, we first re-order its nodes by a breadth-first search, then randomly remove the last $m$ nodes and edges incident to them, where $m$ is randomly selected from $\{0,1,2,3,4,5\}$. The remaining sub-graph is set as the initial state. 

We use Tanimoto similarity with Morgan fingerprint \cite{rogers2010extended} to compute the similarity between the input and the optimized molecules. We optimize the input 800 molecules under 4 different similarity constraints of 0.0, 0.2, 0.4, 0.6, respectively. The success rate under each similarity constraint is reported, which is the percentage of molecules that can be successfully optimized over 800 molecules. Over all successfully optimized molecules, we calculate the mean and standard deviation of the largest property improvement, and similarities between them and their corresponding input molecules.

\textbf{Results.} Results of constrained optimization on two datasets are summarized in Table~\ref{tab:co_jtvae} and Table~\ref{tab:co_graphaf}, respectively. Results show that GraphDF can achieve much higher average property improvement than JT-VAE, GCPN, and MoFlow under 4 similarity constraints, and outperform GraphAF under 3 of 4 similarity constraints. In addition, the average similarities achieved by GraphDF are comparable with all baselines, and unlike JT-VAE or MoFlow, the success rate of GraphDF does not decrease dramatically as the similarity threshold increases. The good performance further proves that GraphDF has a strong ability to search molecules with high property scores in chemical space. We illustrate several optimization examples in Figure~\ref{fig:co}.

\section{Conclusion}

We propose GraphDF, the first molecular graph generation framework using discrete latent variables. GraphDF sequentially map discrete latent variables to graph nodes and edges via invertible modulo shift transforms. The use of discrete latent variables results in more accurate modeling of graph structures and stronger capacity to explore the chemical space. More importantly, GraphDF can avoid expensive computation of Jacobian matrix and eliminate the negative effects of dequantization, which exist in many prior methods. We experimentally demonstrate the advantages of GraphDF and achieve new state-of-the-art results in three molecule generation tasks. The major limitations of GraphDF are relying on BFS node ordering during training and the generation speed is slower than one-shot methods. In the future, we would like to apply GraphDF to other types of graph data and extend it to graph translation problems.


\section*{Acknowledgements}
We thank Chence Shi for his help on providing technical details on the implementation of reinforcement learning method. This work was supported in part by National Science Foundation grant DBI-2028361.


\bibliography{ref}
\bibliographystyle{icml2021}

\clearpage
\appendix

\section{Generation Algorithm}
\label{app:gen_alg}

\begin{algorithm}[!htbp]
  \caption{Generation Algorithm of GraphDF}
  \begin{algorithmic}[1]
  \STATE {\bfseries Input:} GraphDF model, latent distribution $p_{Z_a}$, $p_{Z_b}$, maximum number of nodes $n$, number of node types $k$, number of edge types $c$
  \STATE 
  \STATE Initialize empty graph $G_0$
  \FOR{$i=1$ {\bfseries to} $n$}
  \STATE $z_i\sim p_{Z_a}$
  \STATE $z_i^0=z_i$
  \STATE $H^L=\mbox{R-GCN}(G_{i-1})$
  \STATE $h=\mbox{sum}(H^L)$
  \FOR{$d=1$ {\bfseries to} $D$}
  \STATE $\mu_i^d=\arg\max\mbox{MLP}^d_a(h)$
  \STATE $z_i^d=(z_i^{d-1}+\mu_i^d)\bmod k$
  \ENDFOR
  \STATE $a_i=z_i^D$
  \STATE Add a new node with type $a_i$ to $G_{i-1}$ and set the updated graph as $G_{i-1,1}$
  \FOR{$j=1$ {\bfseries to} $i-1$}
  \REPEAT
  \STATE $z_{ij}\sim p_{Z_b}$
  \STATE $z_{ij}^0=z_{ij}$
  \STATE $H^L=\mbox{R-GCN}(G_{i-1,j})$
  \STATE $h=\mbox{sum}(H^L)$
  \FOR{$d=1$ {\bfseries to} $D$}
  \STATE $\mu_{ij}^d=\arg\max\mbox{MLP}^d_b\left(\mbox{Con}(h,H_i^L,H_j^L)\right)$
  \STATE $z_{ij}^d=(z_{ij}^{d-1}+\mu_{ij}^d)\bmod (c+1)$
  \ENDFOR
  \STATE $b_{ij}=z_{ij}^D$
  \UNTIL{check\_valency($G_{i-1,j}$, $b_{ij}$) is true}
  \STATE Add a new edge with type $b_{ij}$ connecting the node $i$ and $j$ to $G_{i-1,j}$ and set the updated graph as $G_{i-1,j+1}$
  \ENDFOR
  \IF{$G_{i-1,i}$ is not connected}
  \STATE Delete the $i$-th node from $G_{i-1,i}$ and set it as $G_{i}$
  \STATE Output $G_{i}$
  \ENDIF
  \STATE $G_i=G_{i-1,i}$
  \ENDFOR
  \STATE Output $G_{n}$
  \end{algorithmic}
\end{algorithm}

\newpage
\section{Training Algorithm}
\label{app:train_alg}

\begin{algorithm}[!htbp]
  \caption{Generation Algorithm of GraphDF}
  \begin{algorithmic}[1]
  \STATE {\bfseries Input:} Molecular graph dataset $\mathcal{M}$, GraphDF model with trainable parameter $\theta$, latent distribution $p_{Z_a}$, $p_{Z_b}$, number of node types $k$, number of edge types $c$, learning rate $\alpha$, batch size $B$
  \STATE 
  \REPEAT
  \STATE Sample a batch of $B$ molecular graphs $\mathcal{G}$ from $\mathcal{M}$
  \STATE $L=0$
  \FOR{$G\in\mathcal{G}$}
  \STATE Set $n$ as the number of nodes in $G$
  \STATE Find $S_G=(a_1,a_2,b_{21},a_3,\dots)$ by BFS on $G$
  \FOR{$i=1$ {\bfseries to} $n$}
  \STATE $z_i^D=a_i$
  \STATE Set $G_{i-1}$ as the graph formed by all elements previous to $a_i$ in $S_G$, or an empty graph if $i=1$
  \STATE $H^L=\mbox{R-GCN}(G_{i-1})$
  \STATE $h=\mbox{sum}(H^L)$
  \FOR{$d=D$ {\bfseries to} $1$}
  \STATE $\mu_i^d=\arg\max\mbox{MLP}^d_a(h)$
  \STATE $z_i^{d-1}=(z_i^{d}-\mu_i^d)\bmod k$
  \ENDFOR
  \STATE $z_i=z_i^0$
  \STATE $L=L-\log p_{Z_a}(z_i)$
  \FOR{$j=1$ {\bfseries to} $i-1$}
  \STATE $z_{ij}^D=b_{ij}$
  \STATE Set $G_{i-1,j}$ as the graph formed by all elements previous to $b_{ij}$ in $S_G$
  \STATE $H^L=\mbox{R-GCN}(G_{i-1,j})$
  \STATE $h=\mbox{sum}(H^L)$
  \FOR{$d=D$ {\bfseries to} $1$}
  \STATE $\mu_{ij}^d=\arg\max\mbox{MLP}^d_b\left(\mbox{Con}(h,H_i^L,H_j^L)\right)$
  \STATE $z_{ij}^{d-1}=(z_{ij}^{d}-\mu_{ij}^d)\bmod (c+1)$
  \ENDFOR
  \STATE $z_{ij}=z_{ij}^0$
  \STATE $L=L-\log p_{Z_b}(z_{ij})$
  \ENDFOR
  \ENDFOR
  \ENDFOR
  \STATE $L=\frac{L}{B}$
  \STATE $\theta=\theta-\alpha\nabla_\theta L$
  \UNTIL{$\theta$ is converged}
  \STATE Output GraphDF model with parameter $\theta$
  \end{algorithmic}
\end{algorithm}

\newpage
\newpage
\section{Data Information}
\label{app:data}

\textbf{Molecule datasets.} For random generation of molecular graphs, we use three datasets ZINC250K \cite{irwin2012zinc}, QM9 \cite{ramakrishnan2014quantum}, and MOSES \cite{polykovskiy2020moses}. ZINC250K contains around 250K drug-like molecules selected from ZINC, which is a free public chemical library for drug discovery. The maximum number of nodes among all molecules in ZINC250K is 38, and all nodes belong to 9 different types of heavy atoms. QM9 collects around 130K molecules with up to 9 heavy atoms for quantum chemistry research. MOSES provides a benchmarking platform particularly for evaluating molecule generation models, containing 1.9M molecules in total. The information about three molecule datasets are summarized below in Table~\ref{tab:mol_data}. All molecules are transformed to kekulized form before training, that is, removing hydrogen atoms and replacing aromatic bonds by double bonds. Hence, there are three edge types in total, corresponding to single bonds, double bonds and triple bonds in molecules.
\begin{table}[ht]
    \caption{Information of molecule datasets.}
    \label{tab:mol_data}
    \vskip 0.15in
    \begin{center}
    \begin{small}
    \begin{tabular}{lccc}
        \toprule
        Dataset & \makecell[l]{Number of \\ molecules} & \makecell[l]{Maximum num-\\ber of nodes} & \makecell[l]{Number of \\ node types} \\
        \midrule
        ZINC250K & 249,455 & 38 & 9 \\
        QM9 & 133,885 & 9 & 4 \\
        MOSES & 1,936,962 & 30 & 7 \\
        \bottomrule
    \end{tabular}
    \end{small}
    \end{center}
    \vskip -0.1in
\end{table}

\textbf{COMMUNITY-SMALL and EGO-SMALL.} Following GNF \cite{liu2019gnf}, we evaluate GraphDF on two generic graph datasets, COMMUNITY-SMALL and EGO-SMALL. COMMUNITY-SMALL contains 100 synthetic 2-community graphs, and EGO-SMALL has 200 graphs which are small sub-graphs of Citeseer network dataset \cite{sen2008collective}. We calculate the MMD under two cases. One is calculating MMD between the graphs in the dataset and the set of generated 1024 graph. The other is evaluating on selected graphs from generated 1024 graphs with node distribution matching. If there are $N$ graphs in the dataset, the node distribution matching is done by first computing the distribution over node numbers in the dataset, then selecting $N$ graphs from all generated graphs that closely match this distribution. We use the open source code of \citet{you18graphrnn} to do evaluation.

\section{Experiment Details}
\label{app:exp_detail}

\textbf{Random generation.} On ZINC250K, QM9 and MOSES, the GraphDF model is trained with Adam optimizer for 10 epochs, where the fixed learning rate is 0.001 and the batch size is 32. On COMMUNITY-SMALL and EGO-SMALL, the GraphDF model is trained with Adam optimizer for 1000 epochs, where the fixed learning rate is 0.001 and the batch size is 16. As for generation, a widely used strategy for improving generation quality is to apply temperature parameters in prior distribution. For instance, GraphAF \cite{shi2020graphaf} and MoFlow \cite{zang2020moflow} both generate graphs by sampling from a spherical multivariate Gaussian distribution whose standard deviation is multiplied by a tunable temperature parameter $t$. We adopt the similar strategy in our discrete prior distribution. Specifically, for $p_{Z_a}$ with parameters $(\alpha_0,\dots,\alpha_{k-1})$ and $p_{Z_b}$ with parameters $(\beta_0,\dots,\beta_c)$, we will sample discrete latent variables as
\begin{equation}
    \begin{split}
        p_{Z_a}(z_i=s)&=\frac{\exp{(t_1\alpha_s)}}{\sum_{t=0}^{k-1}\exp{(t_1\alpha_t)}},\\
        p_{Z_b}(z_{ij}=s)&=\frac{\exp{(\beta_s/t_2)}}{\sum_{t=0}^{c}\exp{(\beta_t/t_2)}},\\
    \end{split}
\end{equation}
where $t_1,t_2$ are tunable temperature parameters. Note that there is only one node type in COMMUNITY-SMALL and EGO-SMALL, so only $t_2$ is needed. The temperature parameters used for each dataset are listed below in Table~\ref{tab:temp}.
\begin{table}[ht]
    \caption{Temperature parameters for each dataset.}
    \label{tab:temp}
    \vskip 0.15in
    \begin{center}
    \begin{small}
    \begin{tabular}{lcc}
        \toprule
        Dataset & $t_1$ & $t_2$ \\
        \midrule
        ZINC250K & 0.35 & 0.2 \\
        QM9 & 0.35 & 0.23 \\
        MOSES & 0.3 & 0.3 \\
        COMMUNITY-SMALL & n/a & 0.65 \\
        EGO-SMALL & n/a & 0.5 \\
        \bottomrule
    \end{tabular}
    \end{small}
    \end{center}
    \vskip -0.1in
\end{table}

\textbf{Property optimization.} The model is first pretrained on ZINC250K dataset with the same setting of random generation task for 1000 epochs. Then we apply reinforcement learning to fine-tune it for 200 iterations with a learning rate of 0.0001 and a batch size of 8 using Adam optimizer. During generation, we set the temperature parameters of prior distribution as $t_1=0.8,t_2=0.1$.

\textbf{Constrained optimization.} Same as property optimization, GraphDF model is first pretrained on ZINC250K dataset for 1000 epochs and fine-tuned for 200 iteration. We fine-tune the model with a learning rate of 0.0001 and a batch size of 16 using Adam optimizer. During optimization, we set the temperature parameters of prior distribution as $t_1=1.0,t_2=1.0$. Each molecule is optimized for 200 times.


\end{document}